\documentclass[10pt,twocolumn,letterpaper]{article}

\usepackage{cvpr}              % To produce the CAMERA-READY version
\usepackage{amsmath} 
\usepackage{amssymb}
\usepackage{bm}
 
\usepackage[table]{xcolor}
\usepackage{wrapfig} 

\usepackage{pifont}
\usepackage{threeparttable}
\usepackage{framed}
\usepackage{makecell}

\usepackage{amsfonts}
\usepackage{algorithmic}
\usepackage{graphicx}
\usepackage{textcomp}
\usepackage{arydshln}
\usepackage{booktabs}
\usepackage{multirow}
\usepackage{array}
\usepackage{enumitem}
\usepackage[OT1]{fontenc} 
\newcommand{\cmark}{\ding{51}}%
\usepackage{tikz}

\newsavebox{\circleC}
\newsavebox{\circlestar}

\sbox{\circleC}{%
  \begin{tikzpicture}[baseline=-0.75ex, scale=0.4]
    \node[draw, circle, inner sep=0.2pt] {\footnotesize{C}};
  \end{tikzpicture}%
}

\sbox{\circlestar}{%
  \begin{tikzpicture}[baseline=-0.75ex, scale=0.4]
    \node[draw, circle, inner sep=0.2pt] {\scalebox{1.2}{\textasteriskcentered}};
  \end{tikzpicture}%
}

\usepackage{balance}
\usepackage{stfloats}

\usepackage[pagebackref,breaklinks,colorlinks]{hyperref}

\usepackage[capitalize]{cleveref}
\crefname{section}{Sec.}{Secs.}
\Crefname{section}{Section}{Sections}
\Crefname{table}{Table}{Tables}
\crefname{table}{Tab.}{Tabs.}

\definecolor{mygray}{gray}{.9}
\makeatletter
\newcommand{\thickhline}{%
  \noalign {\ifnum 0=`}\fi \hrule height 1pt
  \futurelet \reserved@a \@xhline
}
\makeatother

\usepackage{lipsum} 

\def\cvprPaperID{3141} % *** Enter the CVPR Paper ID here
\def\confName{CVPR}
\def\confYear{2024}

\begin{document}
 
%\title{\!S2VNet: Universal Volumetric Image Segmentation via Slice-to-Volume Propagation\!\!}
\title{Clustering Propagation for Universal Medical Image Segmentation} 

\author{
Yuhang Ding\textsuperscript{1}, Liulei Li\textsuperscript{1}, Wenguan Wang\textsuperscript{2},~~Yi Yang\textsuperscript{2}\\
\small \textsuperscript{1} ReLER, AAII, University of Technology Sydney~~\textsuperscript{2} ReLER, CCAI, Zhejiang University\\
\small\url{https://github.com/dyh127/S2VNet}
}

%\author{
%Yuhang Ding$^{1}$,~
%LiuLei Li$^{1}$,~
%Wenguan Wang$^{2}$,~
%Yi Yang$^{2}$ \\
%$^1$ReLER, University of Technology Sydney \quad
%$^2$CCAI, Zhejiang University \\ 
%\href{https://github.com/dyh127/S2VNet}{\emph{https://github.com/dyh127/S2VNet}}
%}
%\author{Yuhang Ding\\
%% omit the following lines up until the closing ``}''.
%% Additional authors and addresses can be added with ``\and'',
%% just like the second author.
%% To save space, use either the email address or home page, not both
%\and
\maketitle

%%%%%%%%% ABSTRACT
\begin{abstract}
Prominent solutions for medical image segmentation are typically tailored for automatic or interactive setups, posing challenges in facilitating progress achieved in one task to another.$_{\!}$ This$_{\!}$ also$_{\!}$ necessitates$_{\!}$ separate$_{\!}$ models for each task, duplicating both training time and parameters.$_{\!}$ To$_{\!}$ address$_{\!}$ above$_{\!}$ issues,$_{\!}$ we$_{\!}$ introduce$_{\!}$ S2VNet,$_{\!}$ a$_{\!}$ universal$_{\!}$ framework$_{\!}$ that$_{\!}$ leverages$_{\!}$ \textbf{S}lice-\textbf{to}-\textbf{V}olume$_{\!}$ propagation$_{\!}$ to$_{\!}$ unify automatic/interactive segmentation within a single model and one training session. Inspired by clustering-based segmentation techniques, S2VNet makes full use of the slice-wise structure of volumetric data by initializing cluster centers from the cluster$_{\!}$ results$_{\!}$ of$_{\!}$ previous$_{\!}$ slice.$_{\!}$ 
This enables knowledge acquired from prior slices to assist in the segmentation of the current slice, further efficiently bridging the communication between remote slices using mere 2D networks. Moreover, such a framework readily accommodates interactive segmentation with no architectural change, simply by initializing centroids from user inputs. S2VNet distinguishes itself by swift inference speeds and reduced memory consumption compared to prevailing 3D solutions. It can also handle multi-class interactions with each of them serving to initialize different centroids. Experiments on three benchmarks demonstrate S2VNet surpasses task-specified solutions on both automatic/interactive setups.

\end{abstract}
%%%%%%%%% BODY TEXT
\vspace{-3pt}
\section{Introduction}
In the realm of medical imaging, the practice of precisely revealing anatomical or pathological structure changes in a pixel observation holds the promise to substantially advance diagnostic efficiency\!~\cite{wang2022medical}. Depending on the presence of user interactions, it can be categorized into automatic or interactive medical image segmentation (AMIS/IMIS)\!~\cite{pham2000current}, with the latter involving active user engagement (\eg, click, scribble) throughout the segmentation process\!~\cite{wang2018interactive,zhou2023volumetric}. 

Benefiting from the rapid development of deep learning techniques, both AMIS and IMIS have witnessed great progress in their respective field.  
For AMIS, the emerging of seminal work\!~\cite{ronneberger2015u} leads the research efforts towards developing stronger backbones\!~\cite{milletari2016v,valanarasu2203unext,cciccek20163d}, harnessing multi-scale features\!~\cite{kamnitsas2017efficient,zhou2019unet++,srivastava2021msrf} or incorporating attention mechanism\!~\cite{wang2019abdominal,ding2019hierarchical,wang2019volumetric}, \etc. \!Conversely, IMIS centers its primary focus on effec- tively integrating user inputs into segmentation models\!~\cite{wang2018deepigeos,luo2021mideepseg}, yielding remarkable performance. \!Nevertheless, such a tailored paradigm for each 
% individual 
task greatly diffuses the research endeavors, impeding the seamless transfer of advancement made in one task to another due to the fundamental differences in model architecture and training strategy. Moreover, when working with the same dataset, current solutions necessitate the developing of two separate models for AMIS and IMIS, respectively. This results in a duplication in terms of both training time and network parameters.

\begin{figure}[t]
    \vspace{-5pt}
    \begin{center}
        \includegraphics[width=1.\linewidth]{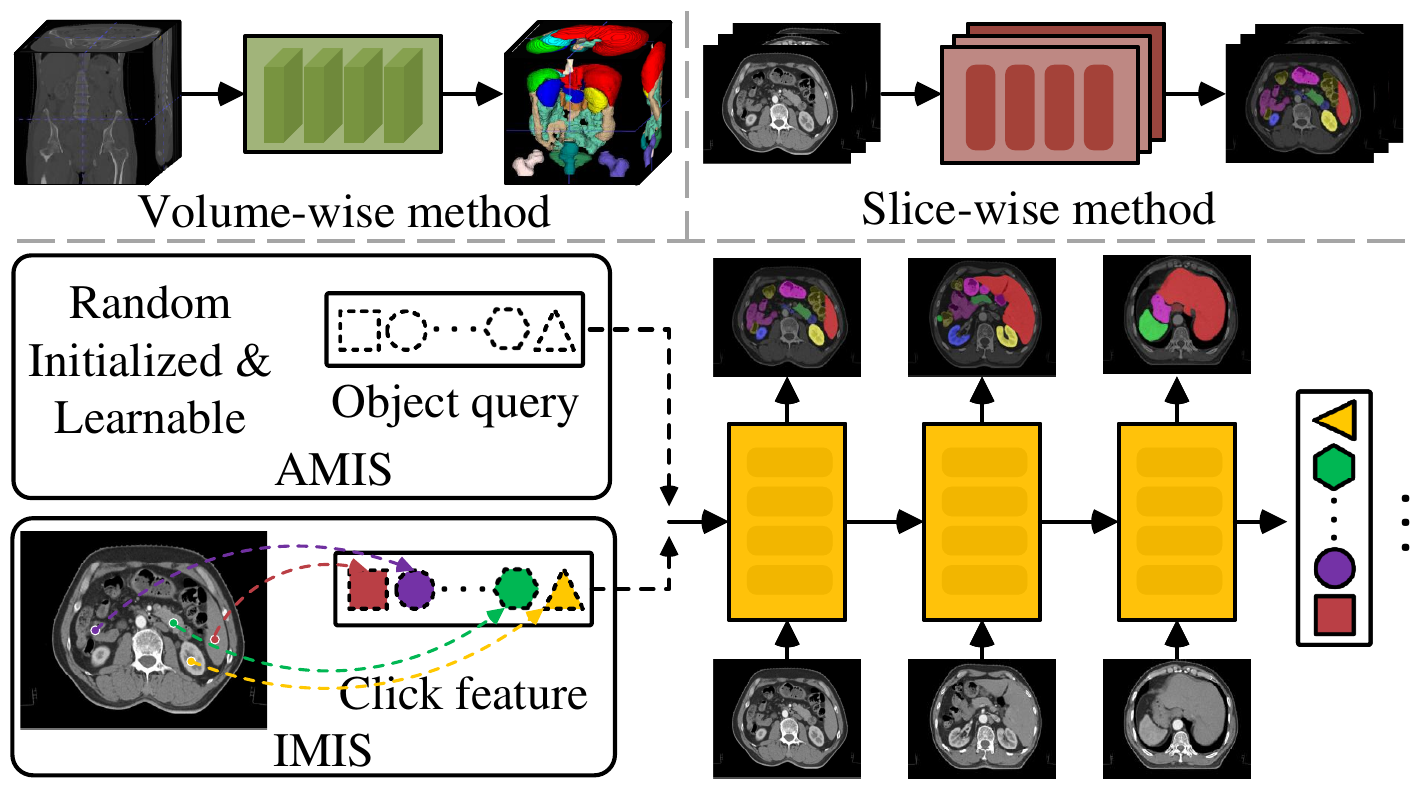}
        \put(-135,95){\small (a)}
        \put(-15,95){\small (b)}
        \put(-15,5){\small (c)}
        \end{center}
    \vspace{-18pt}
    \captionsetup{font=small}
    \caption{\small{(a-b) Existing \textit{volume-wise} and \textit{slice-wise} solutions. (c) Our slice-to-volume solution that bridges distant slices by cluster center propagation and further unifies automatic/interactive segmentation under the same model with 2D segmentation networks.}
    }
    \label{fig:1}
    \vspace{-15pt}
\end{figure}

In this work, we aim to formulate a \textbf{\textit{universal}} segmentation framework capable of
 % efficiently and effectively 
addressing both AMIS and IMIS within \textbf{\textit{one unified model}} and a \textbf{\textit{single training session}}. Towards this, we first initiate a thorough exploration on the limitations commonly observed in current AMIS and IMIS solutions: \textbf{i)} the top-leading approaches for volume segmentation rely heavily on 3D networks which suffer from slow inference\!~\cite{kamnitsas2017efficient} and present significant challenges in deploying on hospital devices that usually exhibit limited parallel computation capabilities, \textbf{ii)} they prove inefficient in bridging remote slices due to the usage of sliding window inference to handle large memory consumption, which further hinders the broadcast of user inputs to entire volumes,
% Here inputs contain only several nearby slices,
\textbf{iii)} current interactive solutions are limited to handle single foreground class, in contrast to automatic approaches which develop rapidly and excel in multi-class segmentation. 

To solve the aforementioned limitations as well as reconcile AMIS and IMIS, we proposed S2VNet. It draws inspiration from the clustering-based image segmentation methods\!~\cite{neven2019instance,yu2022cmt,yu2022k,liang2023clustseg}
that utilize a set of learnable queries as cluster centers to aggregate pixel feature associated with target objects, and update in an iterative manner. This insightful approach prompts us to reformulate volumetric segmentation by utilizing mere$_{\!}$ 2D$_{\!}$ segmentation$_{\!}$ models.$_{\!}$ Specifically,$_{\!}$ it$_{\!}$ is observed that objects in a volume usually manifest identical representation across different slices. This inherent consistency forms the basis for a novel slice-to-volume propagation approach that centroids after comprehensive updates in one slice can be passed forward and serve as the initial values for cluster centers in successive slices, facilitating effortless transfer of knowledge retrieved in prior segmentation process to the next round. This paradigm is simple yet powerful, harnessing both the key principle of clustering-based methodologies and the slice-wise structure of volu- metric$_{\!}$ data.$_{\!}$ Moreover,$_{\!}$ 
% it is imperative to note that 
this$_{\!}$ framework$_{\!}$ is$_{\!}$ readily$_{\!}$ adapted$_{\!}$ to$_{\!}$ IMIS$_{\!}$ without$_{\!}$ architectural$_{\!}$ changes by initializing centroids from backbone features at the position of user inputs,$_{\!}$ which$_{\!}$ clearly$_{\!}$ signify$_{\!}$ intended$_{\!}$ objects.$_{\!}$ Since there would be multiple clicks for a identical object, we further design an adaptive sampling strategy to reweight feature points when given new interactions. Finally, as the current pipeline may be affected by outliers and face decaying awareness to prior cues after rounds of propagation, we devise a recurrent centroid aggregation strategy to collect historic centroids and fuse them into a single vector which introduces nearly no additional cost to deliver a more robust network inference. 

Taking advantage of such slice-to-volume propagation paradigm, S2VNet unveils several compelling facets: \textbf{First}, it seamlessly accommodates AMIS and IMIS into a unified model through a single training process, accomplished by initializing a subset of cluster centers from user inputs while the others are left as random, enabling both automatic and interactive segmentation learning. \ \textbf{Second}, in leveraging of reusing centroids, S2VNet extends user inputs or slice cues throughout the entire volume with 2D networks, contributing to a significant alleviation in computational resource (\ie, 15 times faster inference speed and 48.2\% memory reduction compared to 3D counterparts). \textbf{Third}, S2VNet can simultaneously accept multiple classes of user inputs, with each of them initializing one cluster center. 
This facilitates parallel refinement for multiple instances of \textbf{\textit{different classes}} in a \textbf{\textit{single network forward pass}}, while prior work could only handle one foreground class\!~\cite{wang2018deepigeos,liu2022transforming,zhou2021quality}.  \textbf{Fourth}, given the universal characteristic of S2VNet, we could build a diagnosis system that meets rigorous clinical requirements. Concretely, S2VNet is able to provide coarse segmentation results for multiple lesion/organ classes via AMIS. Physicians can then choose instances of interest and conduct refinement with precise feedback, saving considerable time in the initial search for lesions/organ areas.

% To the best of our knowledge, we are the first that investigate universal segmentation in the field of medical imaging. 
We open a new avenue for medical image segmentation from the universal perceptive, and further provide a feasible solution via clustering-based slice-to-volume propagation.
To comprehensively evaluate our method, we experiment S2VNet on three volumetric datasets, \ie, WORD\!~\cite{luo2022word}, BTCV\!~\cite{landman2015miccai}, and AMOS\!~\cite{ji2022amos}. Our empirical findings substantiate that S2VNet could consistently yield superior performance even compared to the specified solutions for each task, through the utilization of only one single model.

\section{Related Work}
\noindent{\textbf{Volumetric Medical Image Segmentation}} aims to segment organic or pathological structures\!~\cite{ding2021rfnet,ding2021modeling,wang2023genesegnet} in 3D medical images and can be broadly grouped into two categories\!~\cite{zhu20183d}: \textit{slice-wise} and \textit{volume-wise}. The \textit{slice-wise} methods\!~\cite{lalonde2021capsules,wang2019mixed,xu2018quantization,zou2022graph,valanarasu2203unext} usually split 3D images into 2D slices along the z-axis, and then segment each slice separately. Since the proposal of \cite{ronneberger2015u}, there has been a research surge based on the U-shaped architecture\!~\cite{zhou2018unet++,huang2020unet,zhou2019high,gu2019net,valanarasu2021kiu,he2023segmentation,zhu2019ace,lee2020structure,zhang2019net,gu2020net,guo2020spatiotemporal,ding2019hierarchical,wang2021bix,yan2020ms}. Such a paradigm enjoys fast inference but makes no use of the 3D structure of images. In contrast, \textit{volume-wise} methods \cite{milletari2016v,cciccek20163d,ji2020uxnet,yu2020c2fnas,wang2019volumetric,isensee2021nnu} directly process 3D images by extending 2D operations to their 3D counterparts. While capturing spatial context in three dimensions, they are inefficient in establishing meaningful connections between distant regions due to the limited receptive field of CNNs\!~\cite{chen2021transunet}. Recently, efforts have been made to leverage Transformer to capture long-range dependencies\!~\cite{valanarasu2021medical,gao2021utnet,hatamizadeh2022unetr,xie2021cotr,hatamizadeh2021swin,wang2023swinmm,wang2021transbts,zhou2021nnformer}. However, the inputs are still 3D image patches that contains only$_{\!}$ nearby$_{\!}$ slices,$_{\!}$ remaining unable to bridge remote slices.

In this work, we adhere to the \textit{slice-wise} pipeline, but de-\\vise a novel slice-to-volume propagation mechanism characterized by utilizing pixel clustering to facilitate the storage and reusage of knowledge acquired in prior slices. This seamlessly associates predictions across individual slices. Finally, S2VNet combines the strengths of both efficient inference offered by \textit{slice-wise} methods and the effective segmentation achieved in \textit{volume-wise} methods, ultimately leading to accurate and consistent 3D predictions. {The propagation strategy shares a similar spirit to object association in video segmentation~\cite{koner2023instanceformer,heo2023generalized,wang2018semi,li2022locality,li2023unified,zhang2023boosting}. However, these work mainly targets at challenges like fast motion, occlusion, and object reappearing, while S2VNet explicitly modeling object patterns via clustering. This benefits medical segmentation that usually contains no complex contextual cues.}

\noindent{\textbf{Interactive Medical Image Segmentation.}} Though achi- eving promising performance, the above automatic methods still face {challenges~\cite{luo2020adversarial,luo2019taking,luo2021category}} in clinical applications due to the severe biological variation present in medical images\!~\cite{zhou2023volumetric}. In response to this, interactive medical image segmentation (IMIS)\!~\cite{budd2021survey,boykov2001interactive,wang2016slic,feng2021interactive,wang2020uncertainty,berg2019ilastik} is emerging as a practical strategy to improve accuracy by incorporating user interactions, which includes bounding boxes\!~\cite{rajchl2016deepcut,zhou2017fixed}, scribbles\!~\cite{lee2020scribble2label,wang2016slic}, clicks\!~\cite{luo2021mideepseg,wang2018deepigeos,koohbanani2020nuclick}, and extreme points\!~\cite{khan2019extreme}. Moreover, endeavors have been striven to enhance accuracy by emphasizing the effective integration of user interactions, such as extracting informative cue maps\!~\cite{luo2021mideepseg,wang2018deepigeos,marinov2023guiding}, or adapting networks to inference images\!~\cite{sambaturu2021efficient,wang2018interactive}. Recently, alternate research\!~\cite{zhou2021quality,liu2022isegformer,shi2022hybrid,zhou2023volumetric,liu2023exploring} explores interactive segmentation in a mask propagation manner, \ie, wrapping the mask of previous slice according to affinity matrix to predict the next slices. {They differ from S2VNet in two aspects:  \textbf{i) core idea}: they follow the mask wrapping pipeline, while S2VNet passes cluster centers to enable continuous segmentation of targets; \textbf{ii) network architecture}: they need two distinct networks (one providing 2D predictions with user inputs and the other propagating the predictions) to conduct IMIS, and cannot handle AMIS, while S2VNet can tackle both AMIS and IMIS in an unified network.}

\noindent{\textbf{Clustering-Based Image Segmentation.}} Prior to the resurgence of deep learning, clustering stood out as a straightforward yet highly efficient technique for image segmentation\!~\cite{dhanachandra2015image,burney2014k,ray1999determination}.
However, these traditional methods rely heavily on low-level features such as texture or color, limiting their capacity to capture high-level semantics\!~\cite{yu2022k}.
Recent studies have explored the utilization of CNNs to extract feature representations\!~\cite{long2015fully,cheng2022masked,liang2022gmmseg,zhou2022rethinking,wang2021exploring,li2022deep,liang2023logic,li2023logicseg,cheng2023segment,li2023semantic,li2024omg}, with mask predictions are delivered by clustering pixels into semantically coherent segments in a post-processing manner\!~\cite{neven2019instance,kong2018recurrent,feng2023clustering,yin2022proposalcontrast,chen2024rethinking}. 
There has also been a shift towards query-based Transformer approaches\!~\cite{cheng2022masked}. For instance, \cite{yu2022cmt,yu2022k} rethinks the relation between pixel features and object queries by reformulating cross-attention as a clustering solver. On this basis, \cite{liang2023clustseg} introduces a recurrent cross-attention mechanism which unlocks the power of iterative clustering in pixel grouping. Inspired by these work, S2VNet further extends pixel clustering to the continuous segmentation of 3D volumetric images, achieved by propagating clustering results (\ie, centroids) of previous slices to the next. This not only preserves the coherence of segmentation results over the z-axis, but also establishes a robust prior for predicting identical objects. Moreover, it seamlessly and effortlessly adapts IMIS in the same architecture, contributing to the advancement of universal segmentation for volumetric images.

\vspace{-1pt}
\section{Methodology}
\vspace{-2pt}
\subsection{Preliminary: \textbf{\textit{K-Means}} Cross-Attention}
\vspace{-2pt}
Inspired by DETR\!~\cite{carion2020end}, 
contemporary query-based image segmentation methods\!~\cite{cheng2021per,cheng2022masked} typically introduce a set of learnable embeddings as queries to collect pixel features associated with specific objects via \textit{cross-attention}: 
\vspace{-2pt}
\begin{equation} 
\begin{aligned}\label{eq:1}
\hat{\bm{C}}=\bm{C} + {\mathrm{softmax}_{HW}}(\bm{Q}{(\bm{K})}^\top)\bm{V},
\end{aligned}
\vspace{-2pt} 
\end{equation}
where $\bm{C}\!\in\!\mathbb{R}^{N\!\times\!D}$ represents $N$ object queries with dimension size $D$, $\hat{\bm{C}}$ denotes the updated queries, $\bm{Q}\!\in\!\mathbb{R}^{N\!\times\!D}$, $\bm{K}\!\in\!\mathbb{R}^{HW\!\times\!D}$, $\bm{V}\!\in\!\mathbb{R}^{HW\!\times\!D}$ stand for the features for query, key, and value. Here ${\mathrm{softmax}_{HW}}$ means  to conduct $\mathrm{softmax}$ along the spatial dimension of image features, \ie, computing the probability of affiliated to a unique query across all pixels. It is crucial to note that this mechanism involves attending to a substantial number of pixels.
In contrast to above, \cite{yu2022k} devise the \textit{k-means} cross attention:
\vspace{-2pt}
\begin{equation}
\begin{aligned}\label{eq:2}
\hat{\bm{C}}=\bm{C} + {\mathrm{argmax}_{N}}(\bm{Q}{(\bm{K})}^\top)\bm{V}.
\end{aligned} 
\vspace{-2pt}  
\end{equation}
Here, Eq.\!~\ref{eq:2} compels $\bm{Q}$ to query pixel features belonging to a specific object, and subsequently inspect which query embedding within $\bm{C}$ these features correspond to by applying $\mathrm{argmax}$ along the query dimension $N$. Such process is similar to the \textit{k-means}\!~\cite{lloyd1982least} algorithm which proceeds by alternating between the \textit{assignment} and \textit{update} two steps:
\vspace{-4pt}  
\begin{equation}
\begin{aligned}\label{eq:3}
\text{Assignment Step:} \ \ \ \ \hat{\bm{C}}&=\bm{A}\bm{V},\\
\text{Update Step:} \ \ \ \ \hat{\bm{C}}&=\bm{C} + \hat{\bm{C}},
\end{aligned} 
\vspace{-2pt}   
\end{equation}
where $\bm{A}={\mathrm{argmax}_{N}}(\bm{Q}{(\bm{K})}^\top)$ is the assignment matrix (\ie, attention map) where each element indicates whether a pixel feature is assigned to a particular cluster. As a res- ults, following the execution of a succession of Transformer decoder layers composed by \textit{k-means} cross attention, the query embeddings $\bm{C}$ can be regarded as the cluster centers, which adeptly captures the representation of target objects.

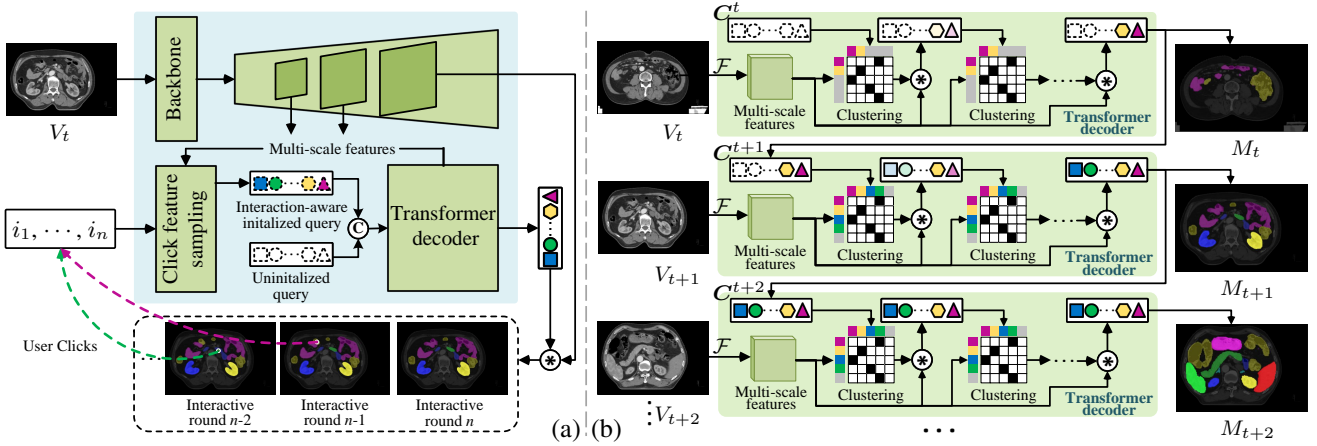
\begin{figure*}[t]
    \vspace{-5pt}
    \begin{center}
        \includegraphics[width=1.\linewidth]{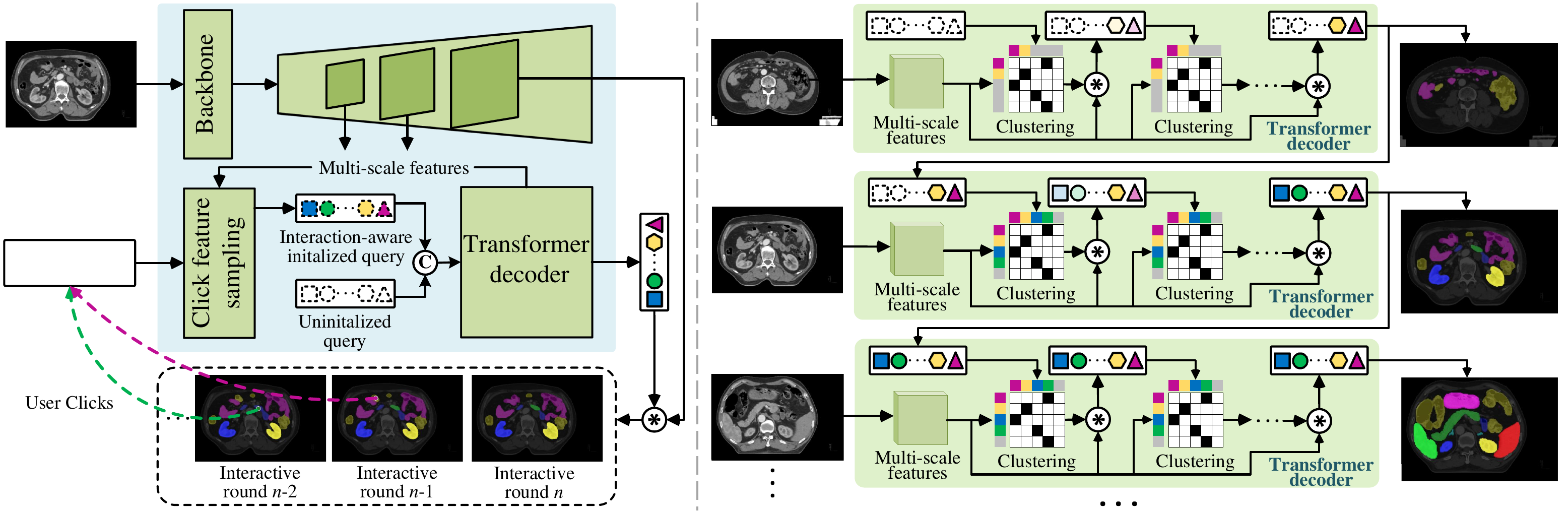}
        \put(-493,76){\small $i_1$, $\cdots$, $i_{n}$}
        \put(-289,-0){(a)}
        \put(-274,-0){(b)}
        \put(-228,32){\footnotesize{$\mathcal{F}$}}
        \put(-228,85){\footnotesize{$\mathcal{F}$}}
        \put(-228,138){\footnotesize{$\mathcal{F}$}}
        \put(-228,52){\footnotesize{$\bm{C}^{t+2}$}}
        \put(-228,105){\footnotesize{$\bm{C}^{t+1}$}}
        \put(-228,158){\footnotesize{$\bm{C}^{t}$}}
         \put(-479,113){\footnotesize{$V_t$}}
        \put(-247,113){\footnotesize{$V_t$}}
        \put(-250,59){\footnotesize{$V_{t+1}$}}
        \put(-250,6){\footnotesize{$V_{t+2}$}} 
        \put(-32,107){\footnotesize{$M_t$}}
        \put(-36,54){\footnotesize{$M_{t+1}$}}
        \put(-36,1){\footnotesize{$M_{t+2}$}} 
        \end{center}
    \vspace{-18pt} 
    \captionsetup{font=small}
    \caption{\small{Our clustering propagation-driven universal segmentation$_{\!}$ framework$_{\!}$ (\S\ref{sec:frame}).\! (a) S2VNet adapts multi-class interactive segmentation and refinement by iteratively initializing cluster centers from user clicks. 
    (b) Our clustering-based slice-to-volume propagation pipeline where centroids are evolved during slice-level segmentation and passed to the next slices. 
    \usebox{\circleC} and \usebox{\circlestar} denote concatenation and dot product.
    }}
    \label{fig:2}
    \vspace{-12pt}
\end{figure*}
 
\subsection{Clustering Propagation-Driven Universal Segmentation Framework}  \label{sec:frame}
\noindent\textbf{Motivation.} Given a volume $V\!\in\!\mathbb{R}^{C\!\times\!H\!\times\!W}$ with a spatial size of $H\!\times\!W$ for $C$ slices, volumetric image segmentation aims to group it into a set of segments with corresponding semantic labels. 
This task is distinguished by the inherent structural property of volumetric image data, \ie, anatomical or pathological regions of interest often$_{\!}$ spanning$_{\!}$ across$_{\!}$ multiple$_{\!}$ consecutive$_{\!}$ slices and exhibiting consistent visual patterns. This property allows the same class of targets in distinct slices to be compressed within a shared object-centric representation.  Given this context, we introduce the clustering-based methodologies into volume segmentation, as shown in Fig.\!~\ref{fig:2}. Specifically, our approach involves extending the dynamic evolution of cluster centers $\bm{C}$ which is originally conducted within the image-level mask decoding process (Fig.\!~\ref{fig:2}\!~(a)) to volume-level by using the same collection of $\bm{C}$ throughout the segmentation for all slices in $V$ (Fig.\!~\ref{fig:2}\!~(b)).   
As such, the separate \textit{slice-wise} segmentation for each individual slice is seamlessly integrated into a coherent segmentation process, and iteratively delivering intermediate output for each slice. 

\noindent\textbf{Slice-to-Volume$_{\!}$ Cluster$_{\!}$ Center$_{\!}$ Propagation.}$_{\!}$ Denoting $\mathcal{F}$ as the feature encoder, $N$ cluster centers $\{{\bm{C}}_n^t\}_{n=1}^N$ are employed to extract the object-centric representation for each class within the given slice $V_t$ by:
\vspace{-2pt}
\begin{equation}
\begin{aligned}\label{eq:4}
\{\hat{\bm{C}}_n^t\}_{n=1}^N=\mathcal{D}(\mathcal{F}(V_t), \{{\bm{C}}_n^t\}_{n=1}^N),
\end{aligned} 
\vspace{-2pt}  
\end{equation}
where $\mathcal{D}$ is the Transformer decoder composed of \textit{k-means} cross attention\!~\cite{yu2022k}. 
In the context of automatic volumetric image segmentation, the segmentation often begins from the first slice along the z-axis of the volume, which typically contains no foreground objects. It is the common case that these foreground objects usually appear in the middle part of the volume. To address the challenge that all cluster centers collect features of the background class and further impose negative impact to the segmentation of subsequent slices, only cluster centers matched with foreground classes will be propagated to the next slice. To achieve this, we perform one-to-one bipartite matching between the mask predictions $\{\hat{Y}_n^t\}_{n=1}^N$ and the ground truth $\{Y_k^t\}_{k=1}^K$ by:
\vspace{-2pt}
\begin{equation}
\begin{aligned}\label{eq:5}
\hat{\theta} = {\arg\min}_{\theta\in\Theta_N} \textstyle \sum_{n=1}^N \mathcal{L}_\text{match}(Y_n, \hat{Y}_{\sigma(n)}).
\end{aligned}  
\vspace{-2pt}  
\end{equation}
Here $\hat{\theta}$ represents the optimal assignment among a permutations of $N$ elements $\theta\!\in\!\Theta_N$.
Based on $\hat{\theta}$, we select cluster centers $\{\hat{\bm{C}}_k^t\}_{k=1}^K$ associated with foreground classes and pass them to the next slice $V_{t+1}$ as the initial values:
\vspace{-2pt}
\begin{equation}
\begin{aligned}\label{eq:6}
\{{\bm{C}}_k^{t+1}\}_{k=1}^K=\{\hat{\bm{C}}_k^t\}_{k=1}^K.
\end{aligned} 
\vspace{-2pt}  
\end{equation}
As such, these object-centric representation could encapsulate the coherent appearances of regions across different slices, fostering a more compact and informative representation for subsequent segmentation and analysis.
Note that during the inference stage, we keep elements in $\{\hat{\bm{C}}_n\}_{n=1}^N$ only if the corresponding class $\{\hat{c}_n^t\}_{n=1}^N$ is not identified as the background class, and pass them to subsequent slices.

\noindent\textbf{Interaction-Aware Cluster Center Initialization.} In prior research\!~\cite{wang2018deepigeos,feng2021interactive,wang2020uncertainty}, the user input is conventionally represented as an binary mask $M\!\in\!\{0,1\}^{H\!\times\!W}$ where the foreground region signifies user guidance. Subsequently, $M$ is combined with gray-scale images as inputs to segmentation networks. Though achieving promising results, such a paradigm suffers from several limitation: 
\textbf{i)} concatenating user inputs with images introduces architectural modifications and disrupts the integration with automatic segmentation into a unified framework, 
and \textbf{ii)} prior methods encounter challenges when accommodating multiple semantic classes, thereby limiting the application to more complex scenarios.
 To tackle above limitations, instead of explicitly incorporating user guidance as the inputs to networks, we harness the clustering-based property of S2VNet. Specifically, denoting  $\{Q_k\}_{k=1}^K=\{(P_k, c_k, t_k)\}_{k=1}^K$ as a set of user inputs where each element $Q_k$ represents a click $P_k$ associated for one semantic class $c_k$ annotated on the slice $V_{t_k}$, we initialize the cluster center $\bm{C}$ from user input by:
\vspace{-4pt}  
\begin{equation} 
\begin{aligned}\label{eq:7} 
\hat{\bm{C}}_k&= \texttt{FFN}(\bm{O}_{k}),\\
\bm{O}_{k}&=\texttt{Sample}(\bm{F}_{t_k}, P_{k}),
\end{aligned} 
\vspace{-2pt}   
\end{equation}
where $\texttt{Sample}$ indicates retrieving the point feature $\bm{O}_{k}$ from backbone features $\bm{F}_{t_k}$ of slice $V_{t_k}$ according to the click position $P_{k}$, and $\texttt{FFN}$ is a simple feed forward network to project $\bm{O}_{k}$ to the same size as $\bm{C}$. $\hat{\bm{C}}_k$ further serves to aggregate pixel features similar to user indicated regions and will be passed to subsequent slices. This realizes user-guided segmentation across the whole volume by leveraging above automatic segmentation pipeline, while introducing no modification$_{\!}$ to$_{\!}$ the$_{\!}$ network$_{\!}$ architecture.$_{\!}$ Moreover, it can accommodate an \textbf{arbitrary number of classes} with each of them serving to initialize one cluster center, perfectly addressing all aforementioned limitations. Notably, extending beyond these benefits, such a centroid initialization-based interactive segmentation strategy offers several additional advantages: \textbf{first},  in contrast to prior work treating user interactions and images equally by concatenating them as inputs which 
can not exercise the guidance ability of interactions to the fullest extent,
our interaction-aware centroid initialization implicitly guarantees predictions always conforming to user highlighted regions and enhances interpretability.  
\textbf{Second}, our method enables unified learning for interactive/automatic segmentation, as the only difference lies in the initial states of centroids. The input data, network architecture, and training objectives remain consistent.

\noindent\textbf{Adaptive Pixel Feature Sampling.} 
Interactive segmentation commonly involves multiple rounds of refinement to improve the precision of previously segmentation results by incorporating newly provided user inputs. 
These iterative refinements yield multiple instances of $Q_k$ associated with the same category label, thus calling for an adaptive strategy to initialize cluster centers for a specific semantic category from multiple user inputs. 
As the latest user input should play more important role in refinement compared to prior clicks, we adopt a weighted sum to combine the pixel feature $\bm{O}_k^r$ sampled from the user input at the current refinement round $r$ with those sampled from prior rounds by:
\vspace{-4pt}  
\begin{equation}  
\begin{aligned}\label{eq:8} 
\hat{\bm{O}}_k^r&= \bm{O}_k^r + \beta^{1}\cdot{\bm{O}}_k^{r-1} + \cdots +\beta^{n}\cdot{\bm{O}}_k^{1},\\
 & = \bm{O}_k^r + \beta\cdot\hat{\bm{O}}_k^{r-1}, 
\end{aligned} 
\vspace{-2pt}   
\end{equation} 
where $\hat{\bm{O}}_k^r$ is the weighted output controlled by the factor $\beta\!\in\![0,1]$. Then $\hat{\bm{O}}_k^r$ at each round of refinement is utilized to initialize a new cluster center, delivering a pair of prediction $\{\hat{\bm{M}}_k^r, \hat{\bm{c}}_k^r\}$ where $\hat{\bm{M}}_k^r$ $\!\in\!\mathbb{R}^{C\!\times\!H\!\times\!W}$ is the binary mask score for all $C$ slices in volume $V$ and $\hat{\bm{c}}_k^r\!\in\!\mathbb{R}^{C}$ is the score for class $c_k$. Consequently, multiple predictions are delivered for each semantic class. To obtain the ultimate output, we first multiply  $\hat{\bm{c}}_k$ with corresponding $\hat{\bm{M}}_k$ and then retrieve the maximum value across all $R$ rounds of predictions by:
\vspace{-4pt}  
\begin{equation} 
\begin{aligned}\label{eq:9} 
\hat{{M}}_k&= \max_{R}(\hat{\bm{M}}_k^0\cdot\hat{\bm{c}}_k^0, \cdots, \hat{\bm{M}}_k^R\cdot\hat{\bm{c}}_k^R).\\
\end{aligned}  
\vspace{-2pt}   
\end{equation}  
It is crucial to emphasize that for all refinement rounds in S2VNet, the pixel features associated with user inputs are sampled from the \textit{\textbf{same backbone features}} which only need to be computed once. This stands in stark contrast to prior work\!~\cite{wang2018deepigeos,feng2021interactive,wang2018interactive} that repetitively combines prior results with image data and conducts a full network pass at each refinement round. This also contributes to accelerated inference and enhances the efficiency of computer-aided diagnosis.

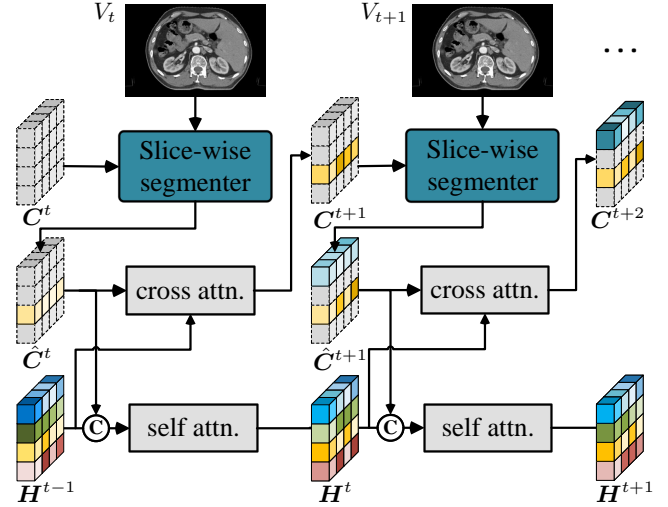
\begin{figure}[t] 
    \vspace{-5pt}  
    \begin{center}
        \includegraphics[width=1.\linewidth]{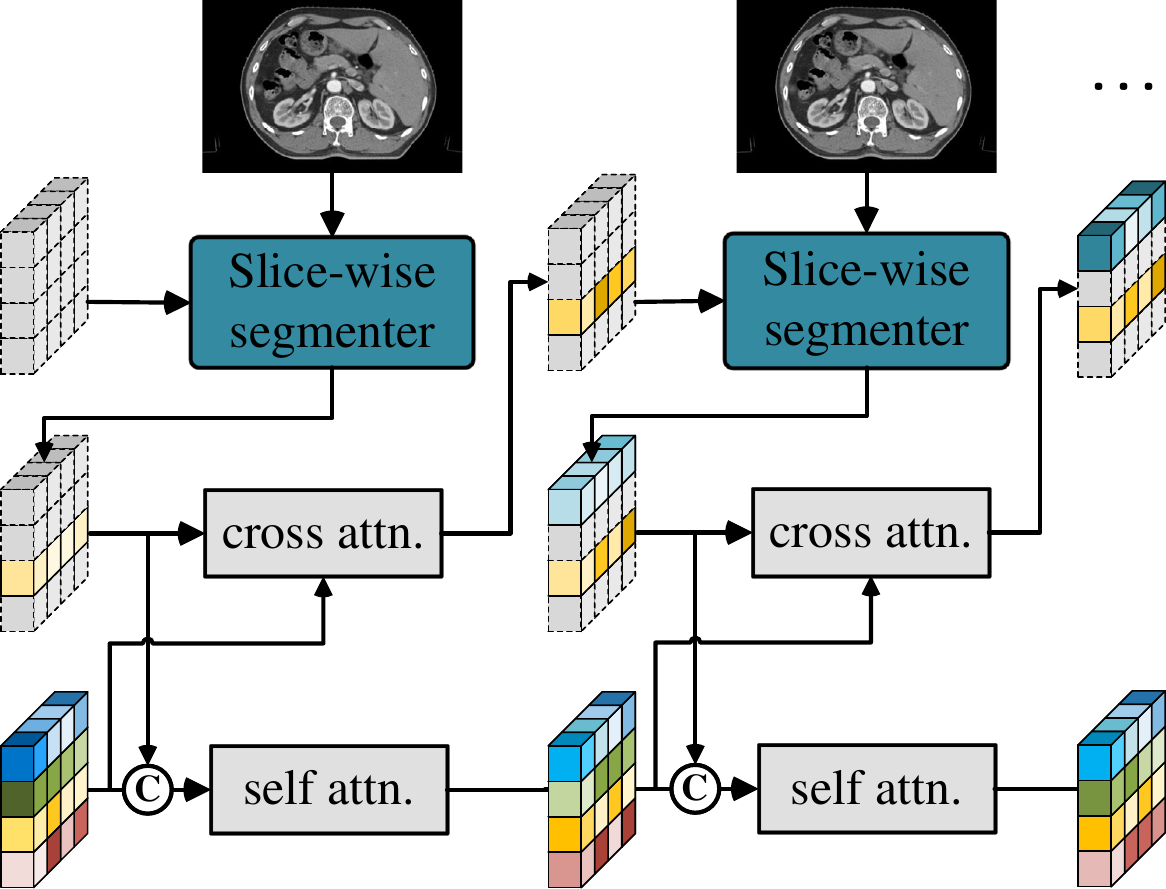}
        \put(-208,175){\small $V_t$}
        \put(-107,175){\small $V_{t+1}$}
        \put(-235,97){\small ${\bm{C}}^t$} 
        \put(-125,96){\small ${\bm{C}}^{t+1}$}
        \put(-20,96){\small ${\bm{C}}^{t+2}$}
        \put(-235,43){\small $\hat{\bm{C}}^t$}
        \put(-125,42){\small $\hat{\bm{C}}^{t+1}$}
        \put(-237,-8){\small $\bm{H}^{t-1}$}
        \put(-123,-8){\small $\bm{H}^{t}$}
        \put(-18,-8){\small $\bm{H}^{t+1}$}
        \end{center}
    \vspace{-18pt}
    \captionsetup{font=small}
    \caption{\small{Illustration of \textbf{recurrent centroid aggregation} (\S\ref{sec:frame}). After clustering within the slice-wise segmentation for each slice, the centroids are recurrently merged with the historic ones to assist in the initialization of centroids belonging to the subsequent slice.}
    }
    \label{fig:3}
    \vspace{-15pt}
\end{figure}

\noindent\textbf{Recurrent Centroid Aggregation.} Though the cluster centers undergo continuous evolution during the mask decoding so as to effectively associate successive slices, they tend to be drifted by outliers such as foreign objects and artifacts commonly encountered in clinical practice\!~\cite{prastawa2004brain}, and lose track of distant structural cues with the slice-wise segmentation process iterates. To deliver a robust inference and re- tain enduring cues of remote slices, we propose to accumu- late historic centroids of each slice and fuse them into a consolidated entity in a recurrent manner. As shown in Fig.\!~\ref{fig:3}, $\bm{H}_{k}^{t-1}$ is denoted as the fused vector, aggregating information from slice $V_0$ to $V_{t-1}$. When given new centroid $\hat{\bm{C}}_{k}^t$ after mask decoding for slice $V_t$, we fuse it with $\bm{H}_{k}^{t-1}$:
\vspace{-4pt}  
\begin{equation} 
\begin{aligned}\label{eq:10} 
\bm{H}_{k}^t = \texttt{FFN}(\texttt{SelfAttn}([\bm{H}_{k}^{t-1}; \hat{\bm{C}}_{k}^t])),
\end{aligned} 
\vspace{-2pt}   
\end{equation} 
where $[;]$ means concatenation. Here the self-attention (\ie, $\texttt{SelfAttn}$) is employed to identify the most relevant regions within the concatenated vector $[\bm{H}_{k}^{t-1}; \hat{\bm{C}}_{k}^t]$, and $\texttt{FFN}$ is subsequently used to project it into the same dimension as $\hat{\bm{C}}_{k}^t$. In this way,  rather than introducing a memory bank which would impose additional GPU memory and computational time overhead, we efficiently store historic structural cues by recurrently merging new centroids into the existing one. Then, when initializing the centroid $\bm{C}_k^{t+2}$ for slice $V_{t+2}$, we incorporate not only the cluster center obtained after mask decoding at slice $V_{t+1}$ (\ie, $\hat{\bm{C}}_k^{t+1}$), but also query the centroids from the previous $t$ slices stored in $ \bm{H}_k^{t}$ by:
\vspace{-4pt}  
\begin{equation} 
\begin{aligned}\label{eq:11} 
\bm{C}_k^{t+2}&= \hat{\bm{C}}_k^{t+1}+\texttt{CrossAttn}(\hat{\bm{C}}_k^{t+1}, \bm{H}_k^{t}).\\
\end{aligned} 
\vspace{-2pt}   
\end{equation} 
Here $\texttt{CrossAttn}$ refers to the standard cross-attention.

\subsection{Implementation details}
\noindent{\textbf{Network Configuration.}} S2VNet is constructed upon the clustering-based image segmenter. Specifically, for the slice-wise segmentation, we adopt Mask2Former\!~\cite{cheng2022masked} and integrate \textit{k-means} cross attention\!~\cite{yu2022k} to replace the standard ones in the Transformer decoder. Other setups remain consistent to the default configuration. The positional encoding in Transformer is reserved to help capture the location of diseases. In order to align S2VNet with the most recent top-leading solutions\!~\cite{hatamizadeh2022unetr,zhou2021nnformer} for medical image segmentation that favor Transformer-based backbones, we employ Swin-B\!~\cite{liu2021swin} for  feature extraction. We empirically set the weighted factor $\beta$ as $0.8$ in adaptive pixel feature sampling.

\noindent{\textbf{Interaction Simulation.}} To evaluate S2VNet under the interactive setup, we opt for click as the primary mode of user interaction which is generally more accessible 
% to a wide range of users 
and can accommodate various input devices like mice, touchscreens, and styluses. Following conventions\!~\cite{xu2016deep,wang2018deepigeos,zhou2021quality}, we adhere to the automatic evaluation pipeline wherein the clicks are simulated based on ground truth and current segmentation results. Specifically, the initial click is
 % strategically 
 sampled near the center of the target object, while subsequent clicks aimed at refinement are generated iteratively from the the most significant error  regions by comparing the current prediction mask with the ground truth. The user clicks comprises both positive and negative ones with the former targeting foreground objects and the latter being applied to background.

\noindent{\textbf{Unified Segmentation Learning.}} To facilitate the slice-to-volume propagation learning, we randomly sample three slices from each volume and use clustering results obtained in previous slice to initialize centroids for the next slice. We designate 20 cluster centers for each semantic class, with each click serving as the trigger to initialize one of them, \ie, allowing up to 20 clicks. Notably, for classes presenting in the inputs, there exists a 50\% probability that the cluster centers are initialized from simulated user clicks, while the left are randomized initialized from empty, so as to enable both automatic and interactive segmentation learning. Following prior work\!~\cite{wang2018deepigeos,zhou2021nnformer,hatamizadeh2022unetr,isensee2021nnu}, the final learning target is the combination of the Cross Entropy loss and Dice loss.

\vspace{-2pt}
\section{Experiment}\label{sec:exp} 
\vspace{-3pt}
\subsection{Experimental Setup}
\vspace{-3pt}
\noindent\textbf{Datasets.}~Our experiments are conducted on three datasets:\!\!
\vspace{1pt}
\begin{itemize}[leftmargin=*]
  \setlength{\itemsep}{0pt}
  \setlength{\parsep}{-2pt}
  \setlength{\parskip}{-0pt}
  \setlength{\leftmargin}{-10pt}
  \vspace{-6pt}
  \item \textbf{WORD}~\cite{luo2022word} is a large-scale real clinical abdomen benchmark, providing high-quality annotations for up to  16 organs in the abdominal region. It contains $100$/$20$/$30$ CT images for \texttt{train}/\texttt{val}/\texttt{test}, respectively.
  \item \textbf{BTCV}~\cite{landman2015miccai} consists of $30$ CT volumes which is divided into $24$ and $6$ volumes for \texttt{train} and \texttt{val}. This dataset provides careful annotation for 13 organs, including 8 of them from Synapse. Following existing work\!~\cite{hatamizadeh2022unetr,zhou2021nnformer}, We report the DSC score on all 13 abdominal organs.
  \item \textbf{AMOS}~\cite{ji2022amos} is a large-scale diverse dataset collected from multiple centers and provides voxel-level annotations for 15 abdominal organs. It covers CT and MRI two modalities with each of them containing $200/100/200$ and $40/20/40$ scans for \texttt{train/val/test}. 
  \vspace{-4pt}
\end{itemize}

\noindent\textbf{Training.}
We train our network for $20$k iterations with a batch size of $8$. The AdamW~\cite{kingma2014adam} optimizer with a initial learning rate $0.0002$ and weight decay $0.02$ is adopted. The learning rate is scheduled following the step policy, \ie, decaying by a factor of $10$ at $14$K and $18$K steps, respectively. A learning rate multiplier of $0.1$ is applied to the backbone which is initialized with ImageNet\!~\cite{deng2009imagenet} pre-trained weights. After adapting the volumetric data in to 2D slices, we employ z-score normalization to rescale image intensities within the range of 0 to 255.
The remaining setups are determined following \cite{isensee2021nnu,hatamizadeh2022unetr,wang2018deepigeos,chen2021transunet,zhou2021quality} for fair comparison. Specifically, for data augmentation,  we use standard large-scale jittering (LSJ) augmentation with a random scaling sampled from range $0.5$ to $1.75$, followed by a fixed-size crop of 512$\times$512 for all datasets\!~\cite{luo2022word,landman2015miccai,ji2022amos}. The random horizontal flipping are also applied to enhance diversity. 
 
\noindent\textbf{Testing.} The inference steps are tailored to optimize the usage of user inputs. Please note that we adopt the identical network architecture and model weight for both two tasks. 
%\vspace{1pt}
\vspace{-1pt}
\begin{itemize}[leftmargin=*]
    \setlength{\itemsep}{0pt}
    \setlength{\parsep}{-2pt}
    \setlength{\parskip}{-0pt}
    \setlength{\leftmargin}{-10pt}
    \vspace{-6pt}%In particular, semantic
    \item \textbf{\textit{Automatic.}} Inference starts from the first slice along the z-axis, proceeding sequentially till the final slice.
    \item \textbf{\textit{Interactive.}} Inference is initiated from  the slice with user inputs and broadcast bidirectionally throughout the entire volume, emphasizing the significance of user interactions. 

    \vspace{-8pt}
\end{itemize}
For fair comparison, we follow prior work\!~\cite{zhou2021nnformer,wang2023swinmm} to use the input resolution of 512$\times$512 for all datasets\!~\cite{luo2022word,landman2015miccai,ji2022amos}.
  
\noindent\textbf{Evaluation Metric.}
Following the standard evaluation protocol\!~\cite{luo2022word,antonelli2022medical,zhou2021nnformer}, We employ Dice Similarity Coefficient (DSC)\!~\cite{dice1945measures}, Hausdorff Distance (HD)\!~\cite{huttenlocher1993comparing} and normalized surface dice (NSD)\!~\cite{ji2022amos} to assess the performance under both automatic and interactive setups. DSC quantifies the overlap between predictions and ground-truths, whereas HD functions for measuring the 3D surface distance between them. To eliminate the impact of outliers, we employ HD95, which captures the 95\% distance of all points in one surface to the other. For NSD,  it scores the category-wise segmentation quality for evaluating precision of boundaries.

\noindent\textbf{Reproducibility.} S2VNet is implemented in PyTorch and trained on four NVIDIA Tesla A100 GPU. Evaluation for all methods is conducted on the same machine. 

\noindent\textbf{IMIS Comparison.} As existing interactive approaches\!~\cite{liao2020iteratively,lee2020scribble2label,zhou2023volumetric,liu2022isegformer} are limited to binary segmentation with single foreground class, we train an independent model for each target class while considering remaining classes as background. To render a more comprehensive comparison, we$_{\!}$ adapt$_{\!}$ the$_{\!}$ top-leading$_{\!}$ automatic$_{\!}$ work$_{\!}$ into$_{\!}$ the$_{\!}$ interactive$_{\!}$ setup$_{\!}$ by concatenating$_{\!}$ user$_{\!}$ clicks$_{\!}$ and$_{\!}$ prior round predictions with image data. Given this substantial workload, we only report performance for several representative classes with relatively lower performance across each dataset.

\begin{table*}[t] 
  \centering
  \small 
     \resizebox{1.00\textwidth}{!}{
  \setlength{\tabcolsep}{2.0pt}
  \setlength{\extrarowheight}{0.5pt}
  %\vspace{-1em}
  \begin{tabular}{l|cc|cccccccccccccccc}
        \thickhline
       & \multicolumn{2}{c|}{Average} &&&&&&&&&&&&&&&& \\  \cline{2-3}
      \multirow{-2}{*}{Methods} 
       &HD95 $\downarrow$
      &DSC $\uparrow$
      &\multirow{-2}{*}{Liv}  
      &\multirow{-2}{*}{Spl}
      &\multirow{-2}{*}{Kid L}
      &\multirow{-2}{*}{Kid R}
      &\multirow{-2}{*}{Sto}
      &\multirow{-2}{*}{Gal}
      &\multirow{-2}{*}{Eso}
      &\multirow{-2}{*}{Pan}
      &\multirow{-2}{*}{Duo}
      &\multirow{-2}{*}{Col}
      &\multirow{-2}{*}{Int}
      &\multirow{-2}{*}{Adr}
      &\multirow{-2}{*}{Rec}
      &\multirow{-2}{*}{Bla}
      &\multirow{-2}{*}{Fem L}
      &\multirow{-2}{*}{Fem R}
      \\ 
      \hline 
      \multicolumn{2}{l}{\textbf{\textit{Automatic Setup}}}\\
      \hline
      UNETR \cite{hatamizadeh2022unetr} &17.34&79.77&94.67&92.85&91.49&91.72&85.56&65.08&67.71&74.79&57.56&74.62&80.40&60.76&74.06&85.42&89.47&90.17 \\
      CoTr \cite{xie2021cotr} &12.83&84.66&95.58&94.90&93.26&93.63&89.99&76.40&74.37&81.02&63.58&84.14&86.39&69.06&80.00&89.27&91.03&91.87 \\
      % DeepLabV3+ \cite{} &9.67&84.91&96.21&94.68&92.01&91.84&91.16&80.05&74.88&82.39&62.81&82.72&85.96&66.82&81.85&90.86&92.01&92.29 \\
      Swin UNETR \cite{hatamizadeh2021swin} &14.24&84.34&96.08&95.32&94.20&94.00&90.32&74.86&76.57&82.60&65.37&84.56&87.37&66.84&79.66&92.05&86.40&83.31 \\
      ESPNet \cite{mehta2018espnet} &15.02&79.92&95.64&93.90&92.24&94.39&87.37&67.19&67.91&75.78&62.03&78.77&72.80&60.55&74.32&78.58&88.24&89.04 \\
      DMFNet \cite{chen20193d} &7.52&85.10&95.96&94.64&94.70&94.96&89.88&79.84&74.10&81.66&66.66&83.51&86.95&66.73&79.62&88.18&91.99&92.55 \\
      % nnUNet \cite{isensee2021nnu} &\\
      LCOVNet \cite{zhao2021lcov} &9.11&85.82&95.89&95.40&95.17&95.78&90.86&78.87&74.55&82.59&\textbf{68.23}&84.22&87.19&69.82&79.99&88.18&92.48&93.23 \\
      SwinMM \cite{wang2023swinmm} &9.35&86.18&96.30&95.46&93.83&94.47&91.43&80.08&76.59&83.60&67.38&\textbf{86.42}&\textbf{88.58}&69.12&80.48&90.56&92.16&92.40 \\
         \arrayrulecolor{gray}\hdashline\arrayrulecolor{black} 
            S2VNet (Ours) &\textbf{4.64}&\textbf{87.36}&\textbf{96.72}&\textbf{96.01}&\textbf{95.84}&\textbf{95.93}&\textbf{91.80}&\textbf{82.96}&\textbf{77.28}&\textbf{85.10}&67.07&86.19&88.46&\textbf{72.40}&\textbf{83.27}&\textbf{91.73}&\textbf{93.30}&\textbf{93.75} \\ 
       \hline
      \multicolumn{2}{l}{\textbf{\textit{Interactive Setup}}}\\
      \hline
            iSegFormer$^\dagger$ \cite{liu2022isegformer}&-&-&-&92.14$^\dagger$&91.07$^\dagger$&93.86$^\dagger$&-&72.01$^\dagger$&73.37$^\dagger$&-&69.52$^\dagger$&-&-&69.91$^\dagger$&48.13$^\dagger$&-&-&- \\

           Mem3D$^\dagger$ \cite{zhou2023volumetric} &-&-&-&94.88$^\dagger$&93.55$^\dagger$&93.96$^\dagger$&-&77.38$^\dagger$&80.61$^\dagger$&-&76.29$^\dagger$&-&-&74.57$^\dagger$&73.37$^\dagger$&-&-&- \\
      SwinMM$^\dagger$ \cite{wang2023swinmm} &-&-&-&95.78$^\dagger$&94.27$^\dagger$&95.11$^\dagger$&-&82.26$^\dagger$&80.33$^\dagger$&-&78.54$^\dagger$&-&-&72.96$^\dagger$&85.12$^\dagger$&-&-&- \\
         \arrayrulecolor{gray}\hdashline\arrayrulecolor{black} 
         S2VNet (Ours) &\textbf{3.28}&\textbf{91.41}&\textbf{96.91}&\textbf{96.37}&\textbf{96.15}&\textbf{96.22}&\textbf{94.79}&\textbf{87.23}&\textbf{86.32}&\textbf{88.51}&\textbf{83.91}&\textbf{90.50}&\textbf{91.17}&\textbf{77.73}&\textbf{90.73}&\textbf{94.35}&\textbf{95.85}&\textbf{95.82} \\
      \thickhline
  \end{tabular}
  }
   \leftline{~~\footnotesize{$^\dagger$: An independent model is trained for each class as prior work can only handle binary segmentation. Given substantial workload, we evaluate 8 classes.}}
  \vspace{-18pt}
  \captionsetup{font=small}
  \caption{\small\textbf{Quantitative segmentation results} with comprehensive scoring for each organ on WORD\!~\cite{luo2022word} \texttt{test}.}
  \label{tab:SOTA WORD}
 \vspace{-10pt} 
\end{table*}

\begin{table}
  \centering
  \small
   \resizebox{1.02\columnwidth}{!}{
  \setlength{\tabcolsep}{3.0pt}
  \setlength{\extrarowheight}{0.5pt}
         \begin{tabular}{l|c|cccccc}
        \thickhline
       & {Avg} \\  
      \multirow{-2}{*}{Method} 
      &DSC
      &\multirow{-2}{*}{Gal}  
      &\multirow{-2}{*}{Eso}
      &\multirow{-2}{*}{IVC}
      &\multirow{-2}{*}{PSV}
      &\multirow{-2}{*}{RAG}
      &\multirow{-2}{*}{LAG}
      \\ 
            \hline
      \multicolumn{2}{l}{\textbf{\textit{Automatic Setup}}}\\
      \hline
      TransUNet \cite{chen2021transunet} &76.72& 59.84& 70.96&77.23 &71.47&65.24&64.06\\
      TransBTS \cite{wang2021transbts} &81.31 &68.38 &75.61  &82.48 &74.21 &67.23 &67.03 \\
      UNETR \cite{hatamizadeh2022unetr} &76.00 &58.23 &71.21  &76.51 &70.37&66.25 &63.04 \\
      Swin-UNETR \cite{hatamizadeh2021swin}  &80.44 &65.37 &75.43  &81.61 &76.30 &68.23 &66.02 \\
      nnFormer \cite{zhou2021nnformer}  &81.62 &65.29 &76.22 & 80.80&75.97 &\textbf{70.20} &66.05\\ 
      % TransUNet \cite{chen2021transunet} &76.72& 59.84& 70.96& 71.38&67.49&65.24&64.06\\
      % TransBTS \cite{wang2021transbts} &81.31&68.38&75.61&74.21 & 76.02 & 67.23 & 67.03 \\ 
      % % CoTr w/o CNN encoder\cite{xie2021cotr} &11.22&64.4&11.49&71.2&9.59&52.3&12.58&69.8 \\
      % % CoTr\cite{xie2021cotr} &9.70&69.3&9.20&74.6&9.45&55.7&10.45&74.8 \\
      % UNETR \cite{hatamizadeh2022unetr} &  76.00 & 58.23 & 71.21 & 70.37 & 66.06 & 66.25 & 63.04\\
      % Swin UNETR \cite{hatamizadeh2021swin} &80.44&65.37 & 75.43 &76.30 &74.52 &68.23& 66.02 \\
      % nnFormer \cite{zhou2021nnformer} &81.62&65.29&76.22&75.97 & 77.87& \textbf{70.20}& 66.05\\
      3D-UX-Net \cite{lee20223d} &80.76& 64.32& 75.17& 80.42&75.39 &69.52&65.77\\
      \arrayrulecolor{gray}\hdashline\arrayrulecolor{black} 
         S2VNet (Ours) &\textbf{83.81}&\textbf{65.63}&\textbf{78.29}&\textbf{84.41}&\textbf{79.77}&68.38&\textbf{72.28} \\
      \hline
      \multicolumn{2}{l}{\textbf{\textit{Interactive Setup}}}\\
      \hline
                 iSegFormer$^\dagger$ \cite{liu2022isegformer} &-&-&69.37$^\dagger$&72.78$^\dagger$ & -&64.40$^\dagger$& 66.89$^\dagger$\\
           Mem3D$^\dagger$ \cite{zhou2023volumetric} &-&-&74.84$^\dagger$&79.52$^\dagger$ & -& 68.45$^\dagger$& 67.88$^\dagger$\\ 
      nnFormer$^\dagger$ \cite{zhou2021nnformer} &-&-&82.47$^\dagger$&83.65$^\dagger$ & -& 70.41$^\dagger$& 67.34$^\dagger$\\
    
      % 63.27 69.37 72.78 78.43 66.89
         \arrayrulecolor{gray}\hdashline\arrayrulecolor{black} 
        S2VNet (Ours) &\textbf{86.11}&\textbf{69.94}  &\textbf{87.92} &\textbf{89.96} &\textbf{81.64} &\textbf{72.23} &\textbf{73.22}\\
      \thickhline
  \end{tabular}
  }
   \leftline{~~\footnotesize{$^\dagger$: An independent model is trained for each target class. }}
  \vspace{-18pt}
  \captionsetup{font=small}
  \caption{\small\textbf{Quantitative segmentation results} on BTCV\!~\cite{landman2015miccai} \texttt{val}.}
  \label{tab:SOTA BTCV}
 \vspace{-13pt} 
\end{table}

\subsection{Comparison to State-of-the-Arts}
\noindent\textbf{WORD\!~\cite{luo2022word}.}
As shown in Table \ref{tab:SOTA WORD}, S2VNet yields remarkable performance on the automatic setup, \ie, surpassing SwinMM\!~\cite{wang2023swinmm} by \textbf{1.18\%} in terms of DSC and outperforming all 3D solutions in terms of HD95 which emphasizes on the coherence of predictions across slices. This demonstrates the  effectiveness of our 2D slice-to-volume propagation strategy in bridging distance cues. Under the interactive setup, S2VNet achieve a \textbf{4.05\%} average improvement in DSC compared to the automatic setup, verifying the superiority of our interaction-aware centroid initialization strategy. Especially, our approach boosts the performance up to \textbf{83.91\%} for the class `Duo.', surpassing both existing interactive and adapted automatic approaches by a large margin.

\noindent\textbf{BTCV\!~\cite{landman2015miccai}.} Table \ref{tab:SOTA BTCV} compares our method against several top-leading approaches on BTCV\!~\cite{landman2015miccai} \texttt{val}.As seen, S2VNet achieves the best performance on both automatic and interactive setups.
In particularly, compared with nnFormer\!~\cite{zhou2021nnformer} which is the previous SOTA, our approach earns \textbf{2.19\%} improvement in terms of averaged DSC score for the automatic setup. 
This indicates that S2VNet can generalize well to different datasets with various challenging scenarios. We also provide detailed scores for six representative organs with poor performance, where S2VNet gives \textbf{2\%$\sim$6\%} performance gain compared to prior work.

\noindent\textbf{AMOS\!~\cite{ji2022amos}.} Table \ref{tab:SOTA AMOS} confirms again the exceptional performance of S2VNet in the segmentation of both CT and MRI images. Specifically, our algorithm achieves an improvement of \textbf{0.52\%}/\textbf{6.41\%} over 3D-UX-Net\!~\cite{lee20223d} in terms of DSC/NSD. Moreover, with the incorporation of interaction-aware query initialization, S2VNet  consistently surpasses existing methods across all modalities and metrics.

\subsection{Qualitative Comparison Result} \label{sec:vis}
\begin{figure*}
    \vspace{-5pt}
    \begin{center}
        \includegraphics[width=1.\textwidth]{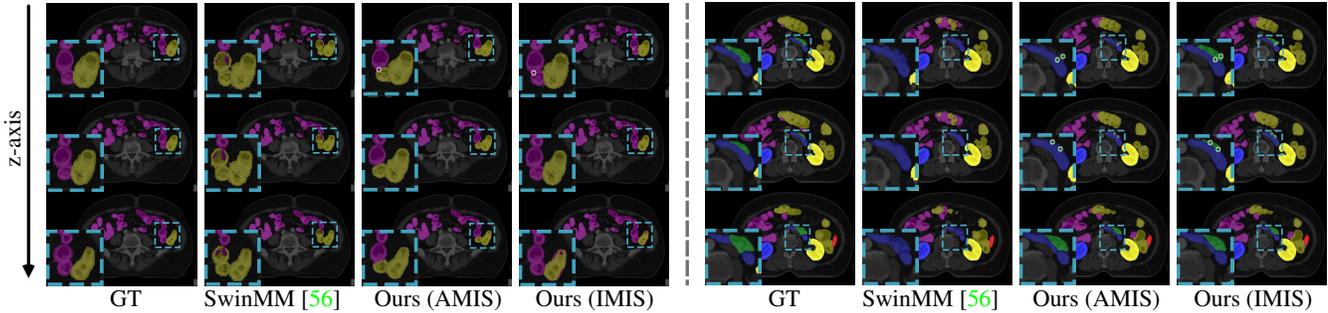}
        \put(-502,48){\small \rotatebox{90}{z-axis}}
        \put(-462,-5){\small GT}
        \put(-426.5,-5){\small SwinMM~\cite{wang2023swinmm}}
        \put(-362,-5){\small Ours (AMIS)}
        \put(-301,-5){\small Ours (IMIS)}
        % \put(-455,8){\small\bfseries GT}
        % \put(-455,8){\small\bfseries GT}
        \put(-213,-5){\small GT}
        \put(-177.5,-5){\small SwinMM~\cite{wang2023swinmm}}
        \put(-113,-5){\small Ours (AMIS)}
        \put(-52,-5){\small Ours (IMIS)}
        \end{center} 
    \vspace{-18pt}
    \captionsetup{font=small}
    \caption{\small{\textbf{Visual comparison results} on WORD\!~\cite{luo2022word} \texttt{test}. See \S\ref{sec:vis} for detailed analysis. }}
    \label{fig:vis}
    \vspace{-12pt}
\end{figure*}

Fig.\!~\ref{fig:vis} depicts visual comparison on WORD\!~\cite{luo2022word} \texttt{test}. As seen, S2VNet yields more accurate results compared to SwinMM\!~\cite{wang2023swinmm}, and the interactive mode can handle various challenging cases with small objects or distortions.
 
\begin{table}
  \centering
  \small
  \resizebox{0.99\columnwidth}{!}{
  \setlength{\tabcolsep}{4.0pt}
  \setlength{\extrarowheight}{0.5pt}
         \begin{tabular}{l|cc|cc|cc}
        \thickhline
       & \multicolumn{2}{c|}{Average} & \multicolumn{2}{c|}{CT} & \multicolumn{2}{c}{MRI} \\   \cline{2-7}
      \multirow{-2}{*}{Method} 
      &DSC$\uparrow$
      &NSD$\uparrow$
      &DSC$\uparrow$
      &NSD$\uparrow$
      &DSC$\uparrow$ 
      &NSD$\uparrow$
      \\ \hline
      \multicolumn{2}{l}{\textbf{\textit{Automatic Setup}}}& & & & &\\
      \hline
      CoTr \cite{xie2021cotr} & 77.31& 67.12 & 77.13 & 64.15 & 77.50 & 70.10\\
      UNETR \cite{hatamizadeh2022unetr} &76.81 & 63.40& 78.33& 61.49&75.30 &65.3 \\
      TransUNet \cite{chen2021transunet}& -& -& 85.05 & 73.86 & - & -  \\
      TransBTS \cite{wang2021transbts}  & -& -& 86.52 & 75.49 & - & -\\
      nnFormer \cite{zhou2021nnformer} & 83.12& 74.07& 85.63& 74.15& 80.60& \textbf{74.00}\\
      Swin UNETR \cite{hatamizadeh2021swin} & 81.04& 70.60& 86.37& 75.32& 75.70& 65.80\\
      3D-UX-Net \cite{lee20223d} & - & - & 87.28 & 76.48 & - & -\\
         \arrayrulecolor{gray}\hdashline\arrayrulecolor{black} 
         S2VNet (Ours) &\textbf{86.22} & \textbf{77.23}&\textbf{87.80} & \textbf{82.89}& \textbf{84.64}&71.57\\
      \hline
      \multicolumn{2}{l}{\textbf{\textit{Interactive Setup}}} & & & & &\\
      \hline
      S2VNet (Ours) &\textbf{88.75} & \textbf{80.94}&\textbf{89.65} &\textbf{85.27} & \textbf{87.84}&\textbf{76.61} \\
      \thickhline
  \end{tabular}
  }
  \vspace{-7pt}
  \captionsetup{font=small}
  \caption{\small\textbf{Quantitative segmentation results} on AMOS\!~\cite{ji2022amos} \texttt{val}.}
  \label{tab:SOTA AMOS}
 \vspace{-5pt} 
\end{table}

\begin{table}
  \centering
  \small
   \resizebox{0.99\columnwidth}{!}{
  \setlength{\tabcolsep}{1.6pt}
  \setlength{\extrarowheight}{0.5pt}
      \begin{tabular}{c|cccc|cc}
        \thickhline
        \multirow{2}{*}{\#} & S2V &Interaction & Adaptive& Recurrent&  \multirow{2}{*}{HD95  $\downarrow$ } &  \multirow{2}{*}{DSC  $\uparrow$ } \\ 
        % \cline{2-5}
        &Propagation &Initialization & Sampling &Aggregation & &    \\
        \hline
     1&& &   & & 16.63 & 78.67 \\ 
        \arrayrulecolor{gray}\hdashline\arrayrulecolor{black}  
        2&\cmark&   & & & 5.03 & 86.19 \\
        3& \cmark&  &  &  \cmark   &4.64 &87.36 \\ 
         \arrayrulecolor{gray}\hdashline\arrayrulecolor{black}  
        4&\cmark & \cmark &   &  & 4.30 & 89.70 \\
        5&\cmark & \cmark &  \cmark &  & 3.79 & 90.64 \\
        6&\cmark & \cmark &  \cmark & \cmark  &\textbf{3.28} &\textbf{91.41}\\
        % \arrayrulecolor{gray}\hdashline\arrayrulecolor{black}       
        \hline
    \end{tabular} 
  }
  \vspace{-7pt}
  \captionsetup{font=small}
  \caption{\small\textbf{Analysis of essential component} on WORD\!~\cite{luo2022word} \texttt{test}.}
  \label{tab:abs_com}
 \vspace{-12pt} 
\end{table}

\subsection{Diagnostic Experiments}
To evaluate the core designs and gain further insights, we conduct a series of ablative studies on WORD\!~\cite{luo2022word} \texttt{test}.

\noindent\textbf{Key Component Analysis.} We first examine the efficacy of each component in Table \ref{tab:abs_com}, where the \textit{row} \#1  indicates directly segmenting each slice using 2D networks without any form of association. Upon the integration of clustering-based slice-to-volume propagation (\ie, \textit{row} \#2), both DSC and HD95 exhibit noteworthy improvement which demonstrate the effectiveness of our design. For interactive segmentation, as seen in \textit{row} \#4, our interaction-aware centroid initialization strategy can bring up to \textbf{3.51\%} performance gains in DSC. With adaptive pixel-feature sampling (\ie, \textit{row} \#5) to fuse different rounds of user interactions, the performance further boosts to \textbf{90.64\%}. Finally, after incorporating recurrent centroid aggregation, S2VNet obtains the best performance on both setups (\ie, \textit{row} \#3 and \#6), underscoring the general compatibility of this module. 

\begin{table}
  \centering
  \small
   \resizebox{0.99\columnwidth}{!}{
  \setlength{\tabcolsep}{3.0pt}
  \setlength{\extrarowheight}{0.5pt}
   \begin{tabular}{l|cc|cc}
      % \hline\thickhline
      % \rowcolor{mygray}
      \thickhline
       Method &Memory (G) $\downarrow$ & Volume Per Minute  $\uparrow$ & HD95 $\downarrow$ & DSC  $\uparrow$ \\ 
      \hline
       CoTr\cite{xie2021cotr}&26 & 0.18 & 12.83 & 84.66  \\
      Swin UNTER\cite{hatamizadeh2021swin}  & 23&0.21 &14.24 &84.34\\
      SwinMM\cite{wang2023swinmm} & 27 & 0.15 &9.35& 86.18 \\
         \arrayrulecolor{gray}\hdashline\arrayrulecolor{black} 
         Baseline &11 & 2.69 &  16.63 & 78.67\\
         \arrayrulecolor{gray}\hdashline\arrayrulecolor{black} 
       S2VNet& \textbf{14}  & \textbf{2.33} &\textbf{4.64} & \textbf{87.36}\\
       \hline
   \end{tabular}
  }
  \vspace{-7pt}
  \captionsetup{font=small} 
  \caption{\small\textbf{Comparison of running efficiency} on WORD\!~\cite{luo2022word} \texttt{test}.\!}
  \label{tab:abs_run}
 \vspace{-9pt} 
\end{table}

\begin{figure}
    \vspace{-5pt}
    \begin{center}
        \includegraphics[width=1.\linewidth]{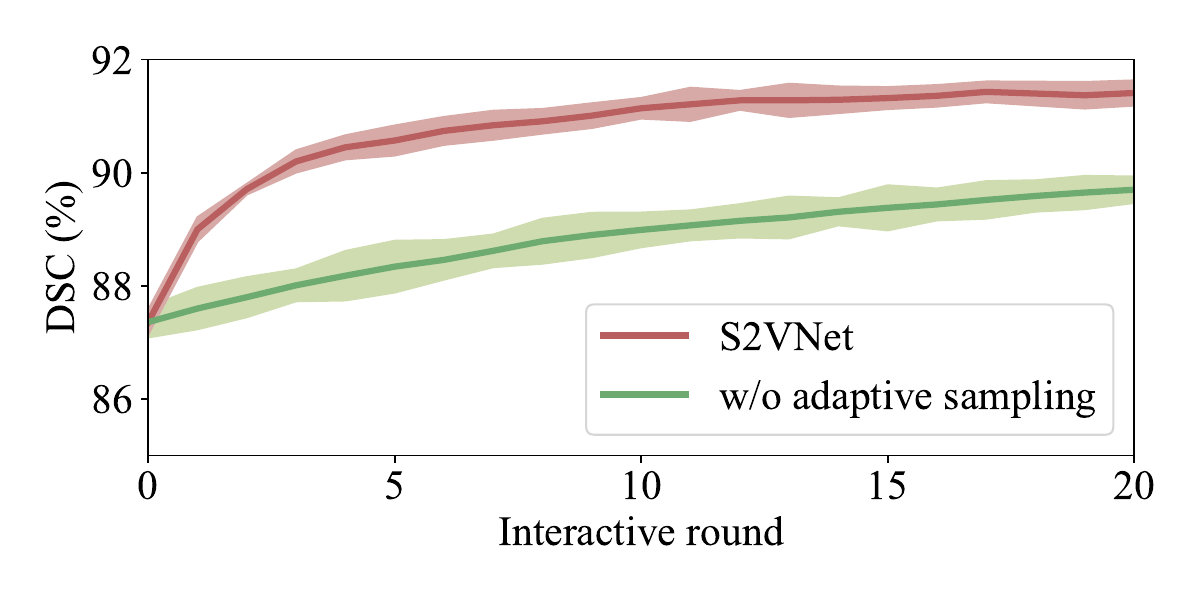}
        \end{center}
    \vspace{-18pt}
    \captionsetup{font=small}
        \vspace{-6pt}
    \caption{\small{\textbf{Convergence analysis} on WORD\!~\cite{luo2022word} \texttt{test}. We report the DSC score with different round of user interactions. }}
    \label{fig:abs_round}
    \vspace{-15pt}
\end{figure}

\noindent\textbf{Run-Time Analysis.} Next we probe the running efficiency of S2VNet during inference. Here `Baseline' represents 2D segmentation network without association. As evidenced in Table \ref{tab:abs_run}, S2VNet achieves nearly \textbf{15} times faster inference speed in terms of FPS and saves \textbf{48.2\%} memory usage compared to the previous state-of-the-art (\ie, SwinMM\!~\cite{wang2023swinmm}). Moreover, our association strategy incurs minor additional cost compared to the baseline method while elevating the performance by an impressive \textbf{8.69\%} in DSC scores. All of the above confirms the urgency of shifting the traditional 3D segmentation paradigm to a more efficient one, with S2VNet providing a pragmatic and effective answer.

\begin{figure}
    \vspace{-5pt}
    \begin{center}
        \includegraphics[width=.49\linewidth]{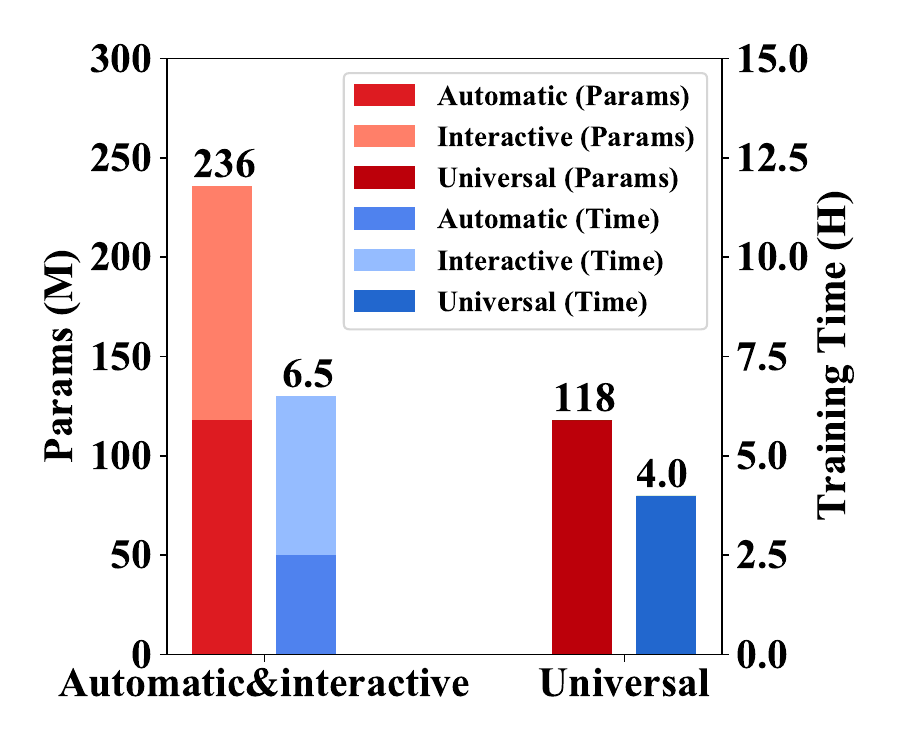}
        \includegraphics[width=.49\linewidth]{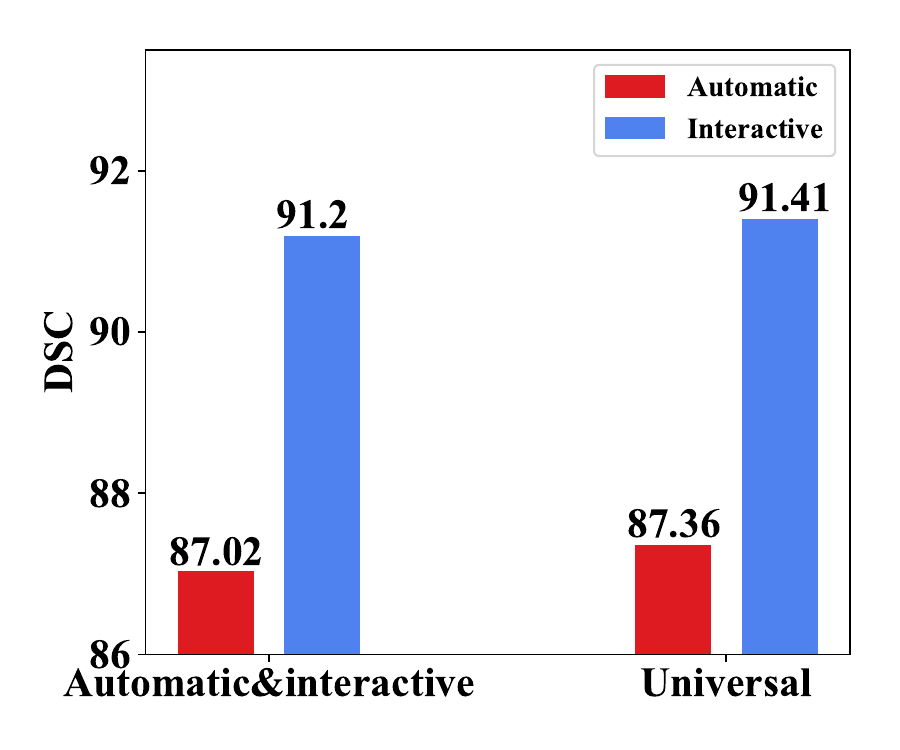}
        % \put(-124,45){\small 15}
        \end{center}
    \vspace{-18pt}
    \captionsetup{font=small}
    \caption{\small\textbf{Analysis of unified training} on WORD\!~\cite{luo2022word} \texttt{test}.}
    \label{fig:abs_unified}
    \vspace{-12pt}
\end{figure}

\noindent\textbf{Convergence Analysis.} We study the correlation between the number of refinement rounds and resulting DSC scores on$_{}$ WORD\!~\cite{luo2022word}$_{\!}$ \texttt{val}$_{\!}$.$_{\!}$ As$_{\!}$ seen in$_{\!}$ Fig.\!~\ref{fig:abs_round},$_{\!}$ 
the$_{\!}$ performance$_{\!}$ of$_{\!}$ S2VNet$_{\!}$ exhibits a stable improvement as the rounds of refinement increase, and consistently outperforms the variant that without adaptive feature sampling to consider interactions in prior rounds. To strike a balance between accuracy and efficiency, we constrain the average refinement rounds to 15 from which there is no significant gain in performance.
\noindent\textbf{Unified Training.} We provide the network parameters and training time comparison between task-specific models for automatic/interactive segmentation and the universal model in Fig.\!~\ref{fig:abs_unified}. As seen, our universal model requires only half of parameters and training times. Furthermore, the performance under such a unified training paradigm even enjoys improvement compared to task-specific training strategies.

\vspace{5pt}
\section{Conclusion}
We present S2VNet, a unified framework to tackle automatic/interaction medical image segmentation in a slice-to-volume propagation manner. It makes use of the clustering-based methods, wherein the knowledge pertaining to targets is compressed within centroids and passed to next slices to produce coherent and robust predictions with merely 2D segmentation networks. On this basis, we realize interactive segmentation by initialize the cluster centers with respect to user guidance. This also facilitates concurrent interaction across multiple classes which overcomes the limitation of prior work confined to binary setups.
Finally, to eliminate the impact of outliers and enhance the awareness to preceding slice cues, a recurrent aggregation approach is proposed to collect historic centroids. All of the above contributes to a flexible solution for volumetric image segmentation characterized by remarkable speed and state-of-the-art accuracy.

\newpage
\newpage
%%%%%%%%% REFERENCES
{\small
\bibliographystyle{ieee_fullname}
\balance
\bibliography{egbib}
}

%%%%%%%%% TITLE STA
% \title{Inter-Video Reconstruction Learning for Temporal Correspondence}
\newpage
\twocolumn[
  \null
  \vskip .375in
  \begin{center}
    {\Large \bfseries Supplemental Material \par}
    %\vspace*{24pt}
%    {
%      \large
%      \lineskip .5em
%      \begin{tabular}[t]{c}
%        Yuhang Ding\textsuperscript{1}, Liulei Li\textsuperscript{1}, Wenguan Wang\textsuperscript{2},~~Yi Yang\textsuperscript{2}\\
%\small \textsuperscript{1} ReLER, AAII, University of Technology Sydney~~\textsuperscript{2} ReLER, CCAI, Zhejiang University\\
%\small\url{https://github.com/dyh127/S2VNet}
%      \end{tabular}
%      \par
%    }
%    \vskip .5em
%    \vspace*{12pt}
\end{center}
]

%%%%%%%%% TITLE END

%\maketitle

% 从这里开始，章节和图形的编号改为字母
\renewcommand{\thesection}{\Alph{section}}
\renewcommand{\thefigure}{S\arabic{figure}}

% 重置计数器，以便从“A”开始
\setcounter{section}{0}
\setcounter{figure}{0}

{
 This document provides additional qualitative results and discussions of S2VNet, which are organized as follows:
 %\begin{itemize}[labelwidth=!,itemsep=0pt,topsep=1pt,parsep=1pt]
    \begin{itemize}
    \setlength{\itemsep}{0pt}
        \item Additional Details (\S\ref{sc:add details})
        \item Discussion (\S\ref{sc:discussion})
 \end{itemize}
}

\section{Additional Details} \label{sc:add details}

\noindent\textbf{Image Preprocessing} %\label{sc:image preprocess}
Following nnUnet\!~\cite{isensee2021nnu}, we firstly clip intensity of CT images into a range of -$175$ to $250$. For MRI images, the clipping is performed using the 0.5 and 99.5 percentiles of the intensity values within each image. Then, we rescale images into the range of $0$ to $255$. 

\begin{figure*}
  \vspace{-5pt}
  \begin{center}
      \includegraphics[width=1.\textwidth]{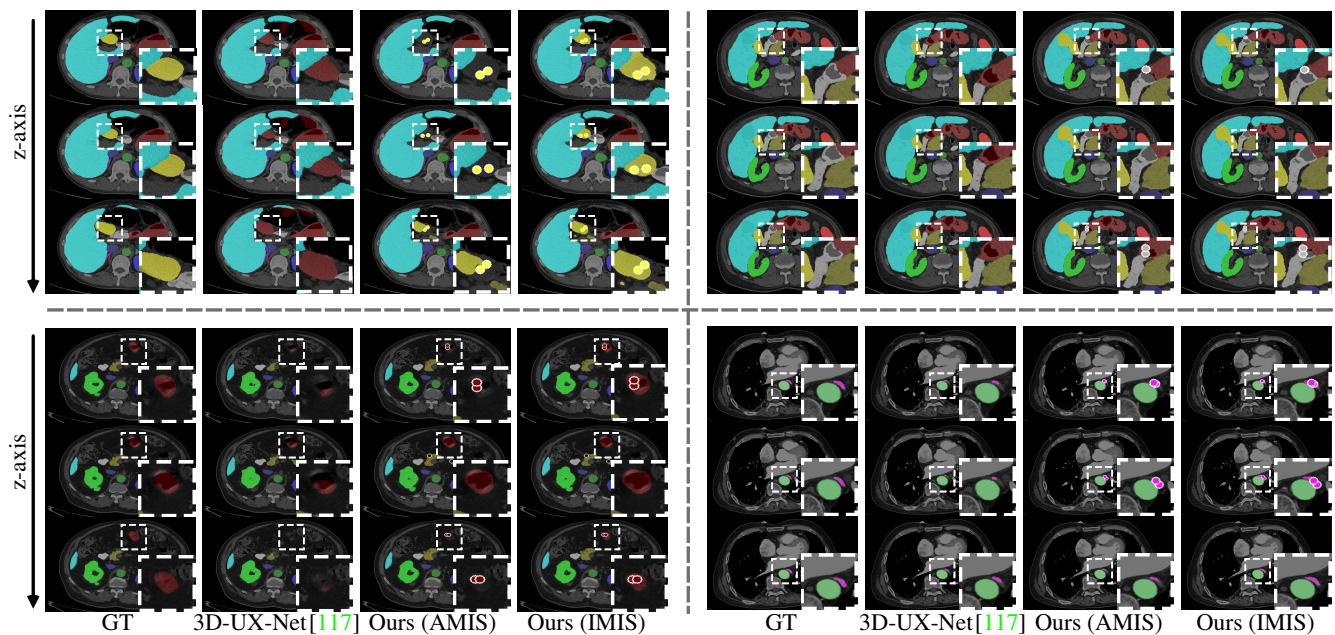}
      \put(-502,48){\small \rotatebox{90}{z-axis}}
      \put(-502,173){\small \rotatebox{90}{z-axis}}
      \put(-467,-5){\small GT}
      \put(-432.5,-5){\small 3D-UX-Net\!~\cite{lee20223d}}
      \put(-366,-5){\small Ours (AMIS)}
      \put(-306,-5){\small Ours (IMIS)}
      \put(-216,-5){\small GT}
      \put(-180.5,-5){\small 3D-UX-Net\!~\cite{lee20223d}}
      \put(-114,-5){\small Ours (AMIS)}
      \put(-53,-5){\small Ours (IMIS)}
      \end{center} 
  \vspace{-18pt}
  \captionsetup{font=small}
  \caption{\small{\textbf{Visual comparison results} on BTCV\!~\cite{landman2015miccai}.}}
  \label{fig:vis btcv}
  \vspace{-5pt}
\end{figure*}

\begin{figure*}
  \vspace{-5pt}
  \begin{center}
      \includegraphics[width=1.\textwidth]{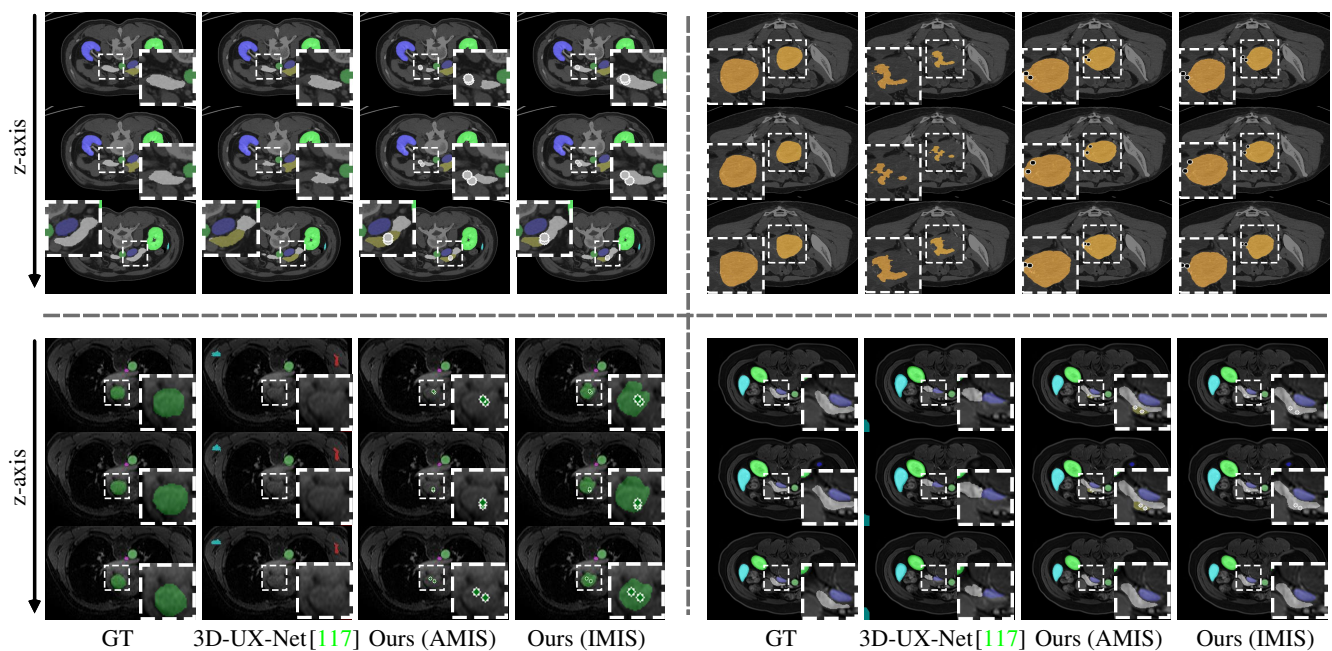}
      \put(-502,48){\small \rotatebox{90}{z-axis}}
      \put(-502,173){\small \rotatebox{90}{z-axis}}
      \put(-467,-5){\small GT}
      \put(-432.5,-5){\small 3D-UX-Net\!~\cite{lee20223d}}
      \put(-366,-5){\small Ours (AMIS)}
      \put(-306,-5){\small Ours (IMIS)}
      \put(-216,-5){\small GT}
      \put(-180.5,-5){\small 3D-UX-Net\!~\cite{lee20223d}}
      \put(-114,-5){\small Ours (AMIS)}
      \put(-53,-5){\small Ours (IMIS)}
      \end{center} 
  \vspace{-18pt}
  \captionsetup{font=small}
  \caption{\small{\textbf{Visual comparison results} on AMOS\!~\cite{ji2022amos} \texttt{val}. \textit{Top}: AMOS CT samples. \textit{Bottom}: AMOS MR samples.}}
  \label{fig:vis amos}
  \vspace{-5pt}
\end{figure*}

\noindent\textbf{Qualitative Results} %\label{sc:vis}
We provide additional visual results on three datasets, \ie, WORD\!~\cite{luo2022word} \texttt{test} in Fig.$_{\!}$~\ref{fig:vis word}, BTCV\!~\cite{landman2015miccai} in Fig.$_{\!}$~\ref{fig:vis btcv}, and AMOS\!~\cite{ji2022amos} \texttt{val} in Fig.$_{\!}$~\ref{fig:vis amos}. 

\noindent\textbf{Failure Cases} %\label{sc:failure vis}
We provide failure cases in Fig.$_{\!}$~\ref{fig:vis visual_failure_cases}. As seen, our method faces challenges in statementing small objects or targets with indistinct boundaries, which is commonly observed in previous method\!~\cite{wang2023swinmm}.

\begin{figure*}[!b]
  \vspace{-5pt}
  \begin{center}
      \includegraphics[width=1.\textwidth]{figure/supp_WORD_visual.pdf}
      \put(-502,48){\small \rotatebox{90}{z-axis}}
      \put(-502,173){\small \rotatebox{90}{z-axis}}
      \put(-467,-5){\small GT}
        \put(-427.5,-5){\small SwinMM~\cite{wang2023swinmm}}
        \put(-366,-5){\small Ours (AMIS)}
        \put(-306,-5){\small Ours (IMIS)}
        \put(-214,-5){\small GT}
        \put(-174.5,-5){\small SwinMM~\cite{wang2023swinmm}}
        \put(-114,-5){\small Ours (AMIS)}
        \put(-53,-5){\small Ours (IMIS)}
        \end{center} 
  \vspace{-18pt}
  \captionsetup{font=small}
  \caption{\small{\textbf{Visual comparison results} on WORD\!~\cite{luo2022word} \texttt{test}.}}
  \label{fig:vis word}
  \vspace{-12pt}
\end{figure*}

\section{Discussion} \label{sc:discussion}
\noindent\textbf{Limitation} %\label{sc:lim}
In comparison to state-of-the-art methods, S2VNet stands out by seamlessly integrating both automatic and interactive medical segmentation, showing notable computational efficiency, and achieving better accuracy in multi-class segmentation. However, S2VNet only addresses predefined classes during automatic segmentation, remaining unable to handle undefined classes. We will explore this direction to improve S2VNet in the future work.

\noindent\textbf{Broader Impact} %\label{sc:soc}
This paper introduces S2VNet, a method with potential applications in various medical contexts, such as the early detection of diseases and the development of personalized treatment plans. Furthermore, the speed and user-friendly nature of S2VNet can not only alleviate the workload of experts but also seek to mitigate the increasing burden faced by the healthcare system as a whole.

\begin{figure*}
  \vspace{-5pt}
  \begin{center}
      \includegraphics[width=1.\textwidth]{figure/visual_failure_cases.pdf}
      \put(-502,48){\small \rotatebox{90}{z-axis}}
      \put(-467,-5){\small GT}
      \put(-427.5,-5){\small SwinMM~\cite{wang2023swinmm}}
        \put(-366,-5){\small Ours (AMIS)}
       \put(-306,-5){\small Ours (IMIS)}
        \put(-214,-5){\small GT}
        \put(-174.5,-5){\small SwinMM~\cite{wang2023swinmm}}
        \put(-114,-5){\small Ours (AMIS)}
       \put(-53,-5){\small Ours (IMIS)}
      \end{center} 
  \vspace{-18pt}
  \captionsetup{font=small}
  \caption{\small{\textbf{Failure cases} on WORD\!~\cite{luo2022word} \texttt{test}.}}
  \label{fig:vis visual_failure_cases}
  \vspace{-12pt}
\end{figure*}

\end{document}